\documentclass{article}

\usepackage{PRIMEarxiv}
\usepackage[utf8]{inputenc}
\usepackage[T1]{fontenc}
\usepackage{tabularx}
\usepackage{cite}
\usepackage{amsmath,amssymb,amsfonts}
\usepackage{graphicx}
\usepackage{booktabs}
\usepackage{multirow}
\usepackage{array}
\usepackage{hyperref}
\usepackage{url}
\hypersetup{hidelinks}
\usepackage{textcomp}
\usepackage{microtype}
\usepackage{orcidlink}
\graphicspath{{media/}}
\newcommand{\IEEEPARstart}[2]{#1#2}
\begin{document}

\title{CFSPMNet: Cross-subject Fourier-guided Spatial-Patch Mamba Network for EEG Motor Imagery Decoding in Stroke Patients}

\author{%
\parbox{0.92\textwidth}{\centering
Xiangkai Wang$^{1}$\,\orcidlink{0009-0001-8600-3274},
Yun Zhao$^{2}$\,\orcidlink{0000-0003-1331-3340}, 
Dongyi He$^{1,3}$\,\orcidlink{0009-0007-9414-0328}, 
Qingling Xia$^{1}$\,\orcidlink{0000-0002-2990-6100},
Gen Li$^{1,4}$, 
Xinlai Xing$^{1}$,
Yuchi Pan$^{1}$\,\orcidlink{0009-0005-2483-6608},
and Bin Jiang$^{1}$\,\orcidlink{0000-0002-7514-0652}\\[0.5em]
\normalfont\small
$^{1}$School of Artificial Intelligence, Chongqing University of Technology, Chongqing 401135, China; Chongqing Key Laboratory of Embodied Intelligence Perception and Autonomous Learning for Humanoid Robots; Key Laboratory of Advanced Equipment Intelligence of the Chongqing Education Commission, Chongqing 401135, China\\
$^{2}$School of Smart Health, Chongqing Polytechnic University of Electronic Technology, Chongqing 401131, China\\
$^{3}$Department of Language Science and Technology, The Hong Kong Polytechnic University, Hung Hom, Hong Kong SAR, China\\
$^{4}$School of Pharmacy and Bioengineering, Chongqing University of Technology, Chongqing 400054, China\\
\texttt{kevinwang@stu.cqut.edu.cn; zhaoyun@cqcet.edu.cn; hedongyi6438@gmail.com}\\
\texttt{qingling@cqut.edu.cn; xingxinlai@cqut.edu.cn; m15938072816@163.com; jb20200132@cqut.edu.cn}\\
\texttt{ligen1990@cqut.edu.cn}\\[0.35em]
Corresponding authors: Yun Zhao (\texttt{zhaoyun@cqcet.edu.cn}) and Bin Jiang (\texttt{jb20200132@cqut.edu.cn})\\
}%
}

\maketitle

\begin{abstract}
Motor imagery electroencephalography (MI-EEG) decoding provides a non-invasive approach for post-stroke rehabilitation; however, its cross-patient application remains challenging due to pathological neural reorganization, which disrupts both task-related EEG dynamics and underlying aperiodic brain activity, altering local neural population excitability, cross-regional coordination, and the global trial-level state context. Consequently, source-learned MI representations become unreliable for unseen patients. To address this, this paper proposes CFSPMNet, a novel cross-patient adaptation framework for post-stroke MI-EEG decoding that explicitly models latent neural-state organization. CFSPMNet integrates a Fourier-Reorganized State Mamba Network (FRSM) with a Shared-Private Prototype Matching (SPPM) module. FRSM treats each trial as a latent physiological token sequence, performing Fourier-domain token-state reorganization to capture trial-specific population-state arrangements and leveraging the Fourier-derived context to guide Mamba state-space propagation. SPPM improves target-domain pseudo-label updating by combining semantic confidence with shared-private physiological consistency,  effectively filtering out confident but physiologically inconsistent target predictions during adaptation. Leave-one-subject-out experiments on two stroke MI-EEG datasets demonstrate that CFSPMNet consistently outperforms representative CNN-, Transformer-, Mamba-, and adaptation-based baselines, achieving average accuracies of 68.23\% on XW-Stroke and 73.33\% on 2019-Stroke, with improvements of 5.63 and 8.25 percentage points over the strongest competitors, respectively. Comprehensive ablation, sensitivity, feature-alignment, pseudo-label selection, and neurophysiological visualization analyses further validate the critical contributions of Fourier-domain token-state reorganization and calibrated pseudo-label updating. These findings highlight that modeling post-stroke MI-EEG as a latent neural-state organization problem can enhance rehabilitation-oriented cross-patient BCI decoding. The implemented code is available at \url{https://github.com/wxk1224/CFSPMNet}.

\end{abstract}

\keywords{
Stroke rehabilitation, brain-computer interface (BCI), electroencephalography (EEG), motor imagery (MI), aperiodic EEG activity, shared-private prototype matching
}

\section{Introduction}
\label{sec:intro}

\IEEEPARstart{S}{troke} is a major neurological cause of persistent motor dysfunction, creating an ongoing need for effective rehabilitation technologies \cite{Valente2026}. Motor imagery (MI) electroencephalography (EEG) provides a non-invasive brain-computer interface (BCI) signal for detecting motor intention when voluntary movement is weak, delayed, or unavailable \cite{11358787}. By translating imagined movements into feedback commands, MI-EEG decoding can support closed-loop rehabilitation involving visual feedback, robotic assistance, or neuromodulatory interventions \cite{JIANG2026109680}. For clinical application, however, a rehabilitation-oriented BCI must generalize to  new patients with minimal recalibration. Therefore, cross-patient post-stroke MI-EEG decoding represents a critical step toward practical neural systems and rehabilitation engineering.

Deep learning-based MI-EEG decoders have advanced this goal through several complementary approaches. CNN-based models, such as EEGNet \cite{Lawhern2018}, IFNet \cite{Wang2023IFNet}, and ShallowConvNet \cite{Schirrmeister2017}, extract local spatiotemporal EEG patterns. Transformer-based models, including DBConformer \cite{Wang2025DBConformer}, EEGConformer \cite{Song2023EEGConformer}, and MSCFormer \cite{Zhao2025MSCFormer}, extend the modeling scope by capturing global token interactions. Additionally, SSM/Mamba-based models, represented by SlimSeiz \cite{Lu2024SlimSeiz}, enable efficient sequence modeling through selective state propagation. Transfer and adaptation methods, such as SSTDA \cite{Chen2025SSTDA}, UA-DANN \cite{Shen2025UADAAN}, and SSAS \cite{Liu2026SSAS}, further reduce inter-patient variability via domain alignment or pseudo-label updating. While these studies offer valuable tools for transferable EEG decoding,  cross-patient variability is still typically addressed as a distributional discrepancy or a feature representation problem. This approach is insufficient for stroke MI-EEG because the recorded signals comprise both structured task-related dynamics and aperiodic background activity, which together reflect the current organization of neural populations.

In post-stroke MI-EEG, the central challenge is more specific. The EEG trial reflects a combination of event-related neural responses, aperiodic electrophysiological activity, and population-level state transitions \cite{Brake2024, McDonnell2026}. Aperiodic activity is closely associated with background cortical excitability, neural gain, and the operating point of local neural populations \cite{Brake2024A, Cross2025The}, while MI-related decoding also depends on how these local states are coordinated across sensorimotor regions and embedded within a global trial-level brain state. Stroke can therefore disrupt the decoding substrate at multiple levels: local population states may become less stable, cross-regional coordination may become less consistently organized, and the global state context of the trial may vary across patients. Under these conditions, the same MI category may correspond to different latent state trajectories even when the task instructions are identical. Consequently, cross-patient transfer is  constrained by disrupted latent neural-state organization; local EEG noise or feature insufficiency alone cannot explain this failure.

To address this issue, this paper proposes CFSPMNet for cross-patient post-stroke MI-EEG decoding. The central concept is that each trial is first  represented as a latent physiological token sequence, which is then reorganized before state-space propagation. Each token represents a compact observation of local neural population activity, while the token trajectory captures how local states interact across regions and evolve into a global trial state. The Fourier transform is employed as a token-organization operator, providing a global complex-basis perspective on this trajectory. The model projects token trajectories onto complex bases, reorganizes their amplitude-phase structure through learnable spectral mixing, and derives a trial-specific context that summarizes the periodic and aperiodic organization of the current state arrangement. This Fourier-derived context is subsequently injected into Mamba state propagation, enabling context-conditioned state-space propagation that adapts to the local, cross-regional, and global organization of each pathological trial.

CFSPMNet further enhances the reliability of target-domain pseudo-supervision. In cross-patient adaptation, relying solely on prediction confidence is insufficient for accepting target samples, as a target trial may be confidently classified despite its aperiodic background and population-state arrangement being inconsistent with the shared physiological organization of the predicted class \cite{LIN2026115150}. Therefore, pseudo-label updating should be constrained by both semantic confidence and shared-private physiological consistency. In the proposed strategy, shared prototypes serve as class-level anchors representing the local-to-global MI state organization learned from source patients; private physiological signatures preserve individual target-sample state characteristics; and class-wise matching tolerance defines the acceptance boundary for pseudo-label inclusion. This calibrated pseudo-label updating reduces the risk of reinforcing confident but physiologically inconsistent target predictions during adaptation.

The main contributions of this paper are summarized as follows:
\begin{enumerate}
\item CFSPMNet is proposed for cross-patient post-stroke MI-EEG decoding by formulating pathological transfer as latent neural-state organization across aperiodic activity and local-to-global population states.
\item FRSM is developed as a Fourier-context-conditioned Mamba encoder to reorganize physiological tokens over complex bases and inject trial-specific Fourier context into state-space propagation.
\item SPPM is designed to jointly leverage semantic confidence and shared-private physiological consistency for gating pseudo-label updates and suppressing physiologically mismatched target supervision.
\end{enumerate}

The remainder of this paper is organized as follows. Section~\ref{sec:method} presents the details of CFSPMNet. Section~\ref{sec:exp} describes the datasets, preprocessing, evaluation protocol, and implementation settings. Section~\ref{sec:results} reports the experimental results and analysis. Section~\ref{sec:conclusion} concludes the paper.

\section{Method}
\label{sec:method}

\begin{figure}[!t]
\centering
\includegraphics[width=\linewidth]{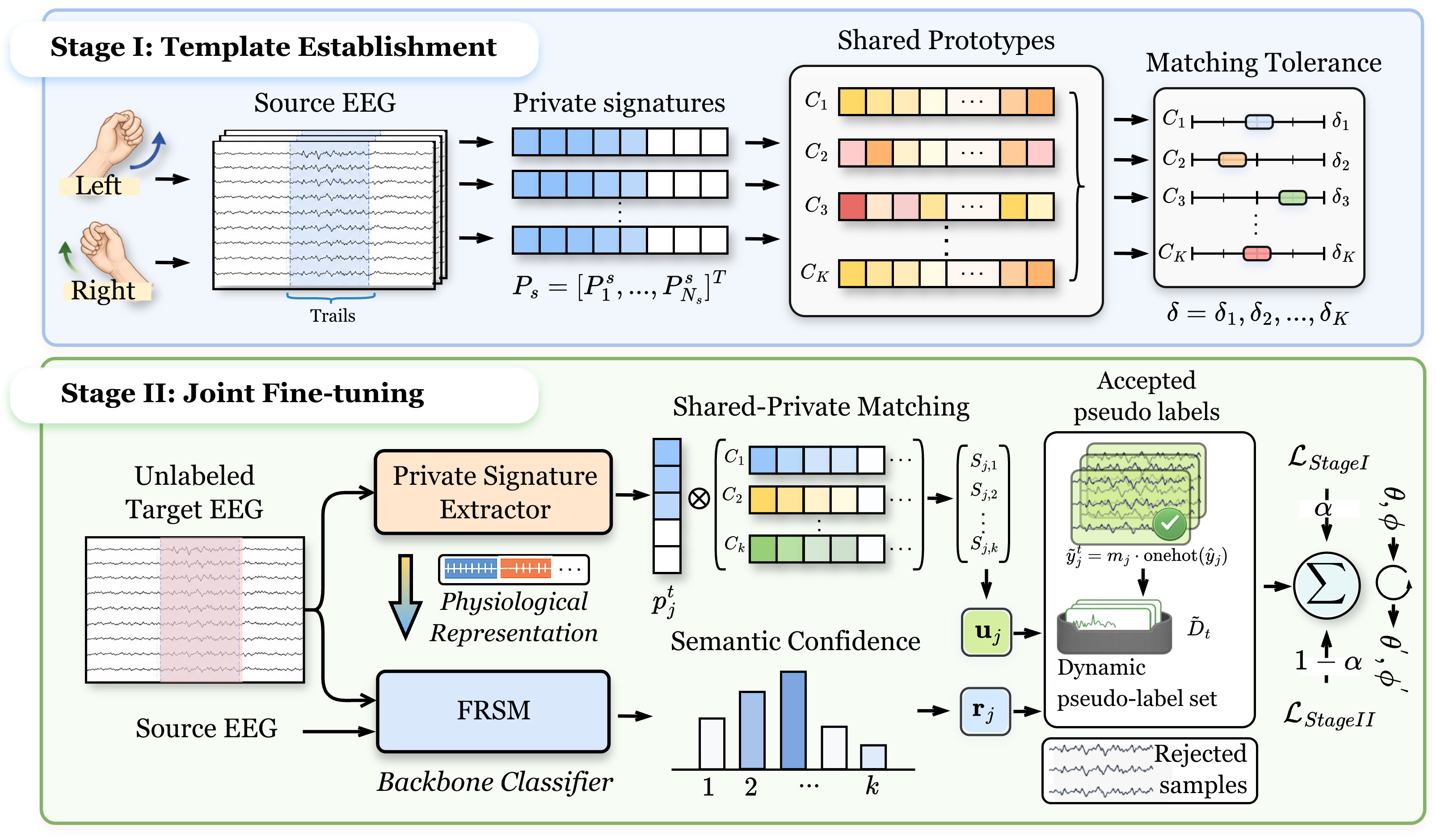}
\caption{Overall framework of CFSPMNet for cross-patient post-stroke MI-EEG decoding.}
\label{fig:framework}
\end{figure}

\begin{figure}[!t]
\centering
\includegraphics[width=\linewidth]{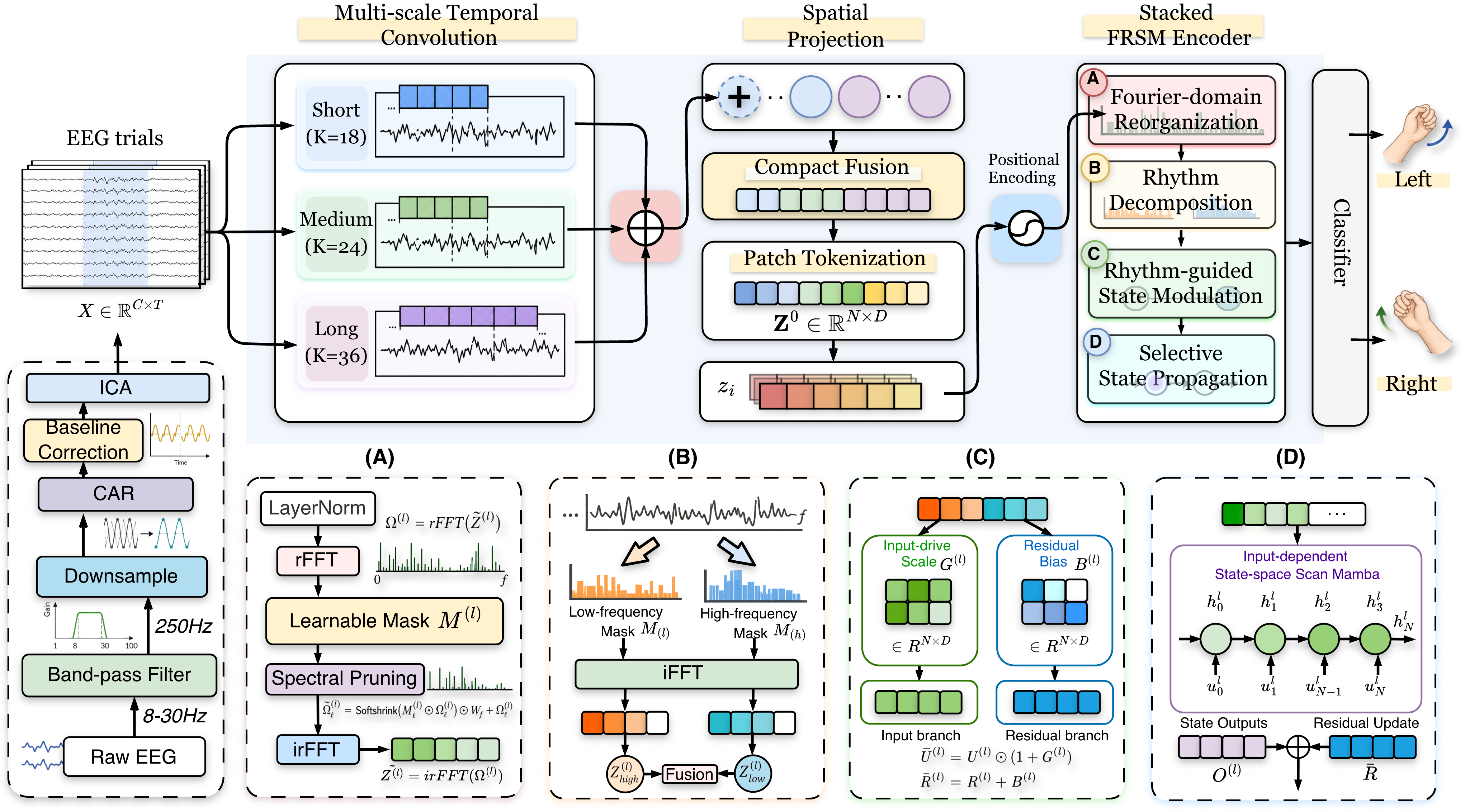}
\caption{Architecture of the Fourier-Reorganized State Mamba Network.}
\label{fig:model}
\end{figure}

\begin{figure}[!t]
\centering
\includegraphics[width=\linewidth]{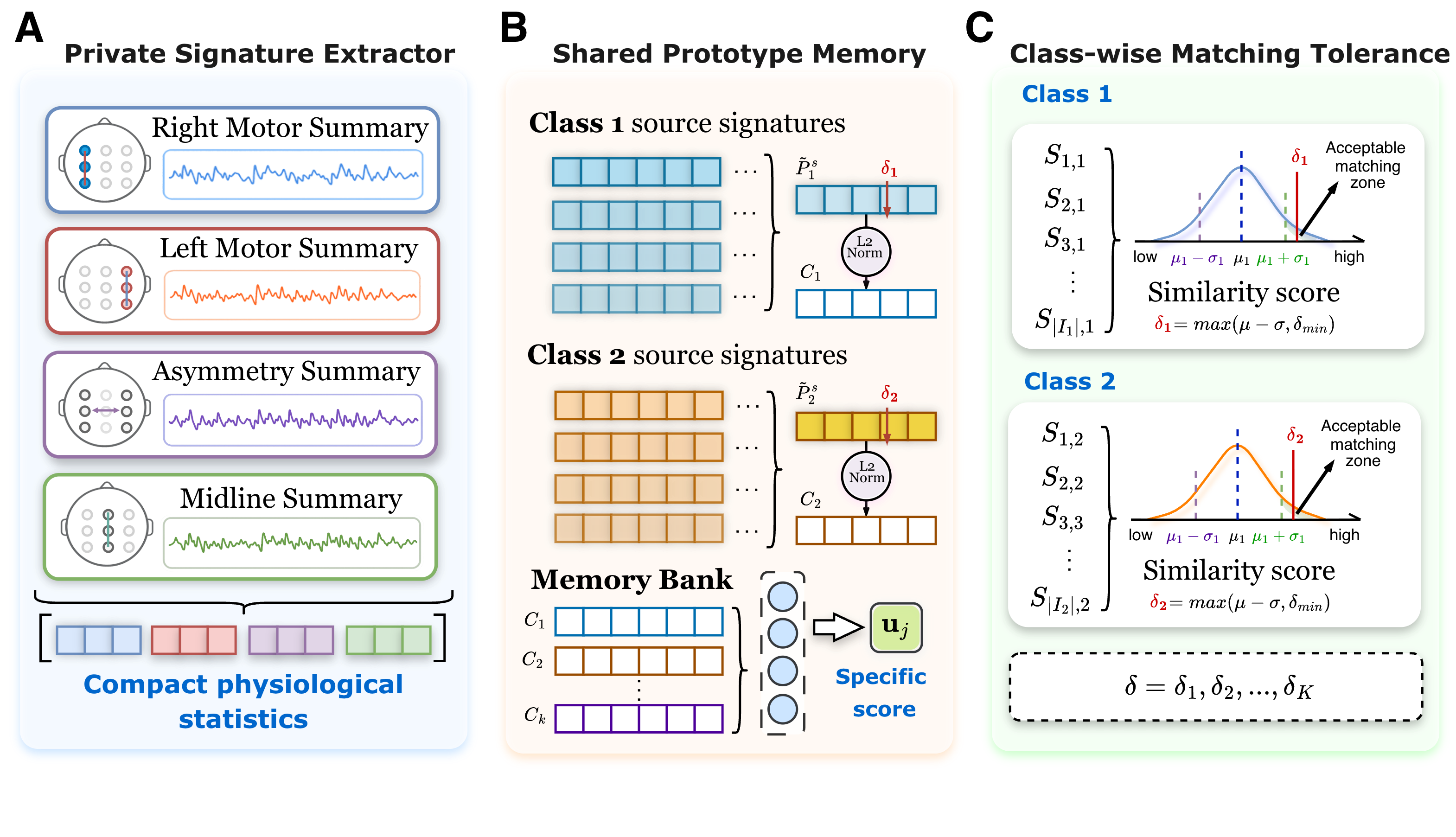}
\caption{Shared-Private Prototype Matching for calibrated target pseudo-label updating.}
\label{fig:sppm}
\end{figure}

CFSPMNet addresses cross-patient post-stroke MI-EEG decoding by coupling Fourier-domain token-state reorganization with calibrated target-domain adaptation, as shown in Figs.~\ref{fig:framework}--\ref{fig:sppm}. The Fourier-Reorganized State Mamba Network (FRSM) converts each EEG trial into spatial-patch physiological tokens, reorganizes their latent state arrangement over complex Fourier bases, and uses the resulting trial-specific context to condition state-space propagation. Shared-Private Prototype Matching (SPPM) constructs source-derived class prototypes in a compact private-signature space and accepts target pseudo-labels only when semantic confidence and shared-private physiological consistency are both satisfied. Training follows two stages: source-supervised initialization with prototype construction, followed by joint source supervision and SPPM-calibrated target pseudo-supervision.

\subsection{Problem Formulation}
\label{subsec:problem_formulation}

Given labeled source trials from multiple patients and unlabeled trials from an unseen target patient, cross-patient decoding learns a transferable function without target labels:
\begin{equation}
\begin{aligned}
\mathcal{D}_s&=\{(X_i^s,y_i^s)\}_{i=1}^{N_s},&
\mathcal{D}_t&=\{X_j^t\}_{j=1}^{N_t},\\
X&\in\mathbb{R}^{C\times T},&
y_i^s&\in\{1,\dots,K\},
\end{aligned}
\end{equation}
where $C$, $T$, and $K$ denote channels, samples, and MI classes. In each LOSO split, one patient is held out as target and all remaining patients form the source domain. The decoder is
\begin{equation}
\begin{aligned}
z&=f_{\theta}(X),\qquad
p(X)=\operatorname{softmax}(g_{\phi}(z)),\\
\hat{y}&=\arg\max_k p_k(X).
\end{aligned}
\end{equation}
Post-stroke transfer is therefore treated as latent-state adaptation: source supervision learns class-discriminative neural-state structure, and unlabeled target trials update the boundary only when their pseudo-labels are accepted into $\mathcal{A}_t$.

\subsection{Fourier-Reorganized State Mamba Network}
\label{subsec:frsm}

FRSM implements $f_{\theta}(\cdot)$ by converting each trial into a token trajectory, treating that trajectory as an observation of local-to-global neural-population states, and conditioning selective state propagation on its Fourier-domain organization. The Fourier representation is used to reorganize token-state arrangement and summarize trial-specific amplitude-phase context, rather than to extract conventional band features.

\subsubsection{Physiological Tokenization}
Given temporal branches $\{\mathcal{T}_m\}_{m=1}^{M}$, spatial mapping $\mathcal{S}(\cdot)$, temporal fusion $\mathcal{F}(\cdot)$, and patch projection $\Pi(\cdot)$, the initial token sequence is
\begin{equation}
\begin{aligned}
F_{\mathrm{temp}}&=\operatorname{Concat}(\mathcal{T}_1(X),\dots,\mathcal{T}_M(X)),\\
F_{\mathrm{emb}}&=\mathcal{F}(\mathcal{S}(F_{\mathrm{temp}})),\\
Z^{(0)}&=\operatorname{PE}\!\left(\sqrt{D}\,\Pi(F_{\mathrm{emb}})\right)\in\mathbb{R}^{L\times D},
\end{aligned}
\end{equation}
where $L$ is token length and $D$ is embedding size.

\subsubsection{Fourier-domain Token-State Reorganization}
For the $\ell$-th encoder block, token states are normalized, transformed along the token dimension, reorganized by a learnable complex mixer, sparsified, and reconstructed as
\begin{equation}
\begin{aligned}
\widetilde{Z}^{(\ell)}&=\operatorname{LN}(Z^{(\ell)}),\qquad
\Omega^{(\ell)}=\operatorname{rFFT}((\widetilde{Z}^{(\ell)})^{\top}),\\
\widehat{\Omega}^{(\ell)}
&=\operatorname{SoftShrink}_{\tau_f}(\mathcal{M}_{f}(\Omega^{(\ell)}))\odot W_f+\Omega^{(\ell)},\\
Z_{\mathrm{enh}}^{(\ell)}
&=\operatorname{iFFT}(\widehat{\Omega}^{(\ell)})+(\widetilde{Z}^{(\ell)})^{\top},
\end{aligned}
\end{equation}
where $\tau_f$ controls sparse shrinkage and $W_f$ denotes learnable filtering. This operation reorganizes the trial-level token-state arrangement while suppressing weak perturbations.

\subsubsection{Context-conditioned State Propagation}
Non-overlapping masks $\{M_b\}_{b=1}^{B}$ with $\sum_{b=1}^{B}M_b=\mathbf{1}$ decompose the reorganized spectrum into complementary states and generate modulation terms:
\begin{equation}
\begin{aligned}
Z_b^{(\ell)}&=\operatorname{iFFT}(\widehat{\Omega}^{(\ell)}\odot M_b)
+(\widetilde{Z}^{(\ell)})^{\top},\\
C^{(\ell)}&=\operatorname{LN}\!\left(\mathcal{H}([Z_1^{(\ell)},\dots,Z_B^{(\ell)}])
+Z_{\mathrm{enh}}^{(\ell)}\right),\\
S^{(\ell)}&=\sigma(W_s C^{(\ell)}),\qquad
B^{(\ell)}=W_b C^{(\ell)} .
\end{aligned}
\end{equation}
Here, $C^{(\ell)}$ is the Fourier-derived context, $S^{(\ell)}$ scales token states, and $B^{(\ell)}$ provides residual bias. The Mamba branch is then conditioned by
\begin{equation}
\begin{aligned}
\relax[U^{(\ell)},R^{(\ell)}]&=\operatorname{Split}(W_{\mathrm{in}}\widetilde{Z}^{(\ell)}),\\
\bar{U}^{(\ell)}&=U^{(\ell)}\odot(1+S^{(\ell)}),\qquad
\bar{R}^{(\ell)}=R^{(\ell)}+B^{(\ell)},\\
h_t^{(\ell)}&=\bar{A}_t^{(\ell)}h_{t-1}^{(\ell)}
+\bar{B}_t^{(\ell)}u_t^{(\ell)},\\
o_t^{(\ell)}&=\bar{C}_t^{(\ell)}h_t^{(\ell)}
+D^{(\ell)}\odot u_t^{(\ell)},\\
z_t^{(\ell,\ast)}&=W_{\mathrm{out}}^{(\ell)}
\!\left(o_t^{(\ell)}\odot\operatorname{SiLU}(\bar{r}_t^{(\ell)})\right),\\
Z^{(\ell+1)}&=Z^{(\ell)}+\operatorname{Drop}(Z^{(\ell,\ast)}),
\end{aligned}
\end{equation}
where $\bar{A}_t^{(\ell)}$, $\bar{B}_t^{(\ell)}$, and $\bar{C}_t^{(\ell)}$ are input-dependent selective parameters and lower-case variables denote token-wise states. After $L$ blocks, classification is performed by
\begin{equation}
p(X)=\operatorname{softmax}\!\left(
g_{\phi}(\operatorname{Flatten}(\operatorname{LN}(Z^{(L)})))\right).
\end{equation}

\subsection{Shared-Private Prototype Matching}
\label{subsec:sppm}

SPPM improves pseudo-label reliability by separating transferable class structure from patient-specific physiological signatures. Source trials define class prototypes and matching tolerances, and a target trial is used only when its prediction is semantically confident and physiologically consistent with the predicted shared class.

\subsubsection{Shared Prototype Memory}
Let $\Psi(\cdot)$ be a private-signature extractor and $\operatorname{sim}(\cdot,\cdot)$ be cosine similarity. For source signatures $\rho_i^s=\Psi(X_i^s)$ with $\|\rho_i^s\|_2=1$, SPPM builds
\begin{equation}
\begin{aligned}
\mathcal{I}_k&=\{i\,|\,y_i^s=k\},\\
c_k&=\operatorname{Norm}_{2}\!\left(
\frac{1}{|\mathcal{I}_k|}\sum_{i\in\mathcal{I}_k}\rho_i^s\right),\\
\mu_k&=\frac{1}{|\mathcal{I}_k|}
\sum_{i\in\mathcal{I}_k}\operatorname{sim}(\rho_i^s,c_k),\\
\sigma_k&=\operatorname{Std}\{\operatorname{sim}(\rho_i^s,c_k)\}_{i\in\mathcal{I}_k},\\
\delta_k&=\max(\delta_{\min},\mu_k-\sigma_k),\qquad
\mathcal{C}=\{(c_k,\delta_k)\}_{k=1}^{K}.
\end{aligned}
\end{equation}
The tolerance $\delta_k$ prevents target updating from being driven by confident predictions outside the source-derived physiological class envelope.

\subsubsection{Shared-Private Pseudo-label Calibration}
In this work, $\Psi(\cdot)$ is implemented as a compact sensorimotor channel summary over fixed channel groups $\{\mathcal{G}_q\}_{q=1}^{Q}$:
\begin{equation}
\begin{aligned}
a_{j,q}&=\operatorname{Avg}_{c\in\mathcal{G}_q}X_{j,c},\qquad
d_j=|a_{j,1}-a_{j,2}|,\\
\rho_j&=\operatorname{Norm}_{2}([a_{j,1},\dots,a_{j,Q},d_j]),\\
s_{j,k}&=\operatorname{sim}(\rho_j,c_k).
\end{aligned}
\end{equation}
For target trial $X_j^t$, classifier confidence and prototype consistency are combined as
\begin{equation}
\begin{aligned}
q_j&=p(X_j^t),\qquad
\hat{y}_j=\arg\max_k q_{j,k},\qquad
r_j=\max_k q_{j,k},\\
u_j&=s_{j,\hat{y}_j},\qquad
m_j=\mathbb{I}[r_j\ge\tau_p]\,\mathbb{I}[u_j\ge\delta_{\hat{y}_j}],\\
\tilde{y}_j^t&=
\begin{cases}
\operatorname{onehot}(\hat{y}_j), & m_j=1,\\
\mathbf{0}, & m_j=0,
\end{cases}
\qquad
\mathcal{A}_t=\{j\,|\,m_j=1\}.
\end{aligned}
\end{equation}
SPPM refreshes $q_j$, $u_j$, $\tilde{y}_j^t$, and $\mathcal{A}_t$ during adaptation, producing calibrated pseudo-label updating instead of confidence-only target supervision.

\subsection{Two-stage Optimization Strategy}
CFSPMNet is optimized in two stages. Stage I trains the encoder and classifier on labeled source trials and constructs $\mathcal{C}$; Stage II recomputes target probabilities, private-signature scores, and accepted pseudo-labels, then updates
\begin{equation}
\begin{aligned}
\mathcal{L}_{\mathrm{src}}&=\frac{1}{N_s}\sum_{i=1}^{N_s}
\ell\!\left(p(X_i^s),y_i^s\right),\\
\mathcal{L}_{\mathrm{tgt}}&=\frac{1}{N_t}\sum_{j=1}^{N_t}
m_j\,\ell\!\left(p(X_j^t),\tilde{y}_j^t\right),\\
\mathcal{L}&=\alpha\mathcal{L}_{\mathrm{src}}
+(1-\alpha)\mathcal{L}_{\mathrm{tgt}},
\end{aligned}
\end{equation}
where $\ell(\cdot,\cdot)$ is cross-entropy and $\alpha$ balances source supervision with calibrated target pseudo-supervision. This schedule stabilizes source class structure before target decision-boundary adaptation.

\section{Experiments}
\label{sec:exp}
\subsection{Datasets}
Before dataset-specific processing, all EEG signals were uniformly preprocessed to preserve MI-related rhythmic activity and suppress non-neural artifacts. The raw EEG was band-pass filtered between 8 and 30 Hz, downsampled to 250 Hz, re-referenced using common average referencing, baseline corrected, and further denoised by ICA-based artifact removal.

\textbf{XW-Stroke:} The XW-Stroke dataset\cite{Liu2024XWStroke} was collected at Xuanwu Hospital, Capital Medical University.  The original dataset contains EEG recordings from 50 acute ischemic stroke patients, including 39 males and 11 females. All subjects performed left-hand and right-hand MI tasks, and EEG was recorded with 30 channels under the international 10–20 system at 500 Hz. To reduce patient-level heterogeneity that was not directly aligned with the cross-subject decoding objective, a clinically defined patient subcohort was further selected according to predefined inclusion criteria: right-handedness, first-ever stroke, NIHSS score below 10, and disease duration below 10 years. These criteria were determined solely by clinical attributes and were applied before model development, independent of training behavior, classification accuracy, or any other experimental outcome. The final cohort included 24 patients, indexed as 2, 5, 8, 9, 11, 12, 14, 17, 21, 23, 24, 26, 27, 28, 30, 32, 33, 37, 38, 43, 44, 47, 49, and 50.

\textbf{2019-Stroke:} The 2019-Stroke dataset was released by Jia \textit{et al.}\cite{Jia2019}. It contains EEG recordings from 15 stroke patients, each performing left-hand and right-hand MI tasks, with both paretic-side and non-paretic-side imagination included in the original acquisition setting. EEG was recorded with 63 channels under the international 10–10 system at 512 Hz. All publicly available subjects were used without additional subject-level exclusion.

\begin{table}[!t]
\caption{Dataset protocols and model input specifications.}
\label{tab:datasets}
\centering
\small
\begin{tabular}{l c l c c c c}
\toprule
\textbf{Dataset} &\textbf{Subjects} &\textbf{System} & \textbf{Channels} & \textbf{Sampling rate} & \textbf{Trials} & \textbf{Input samples} \\
\midrule
XW-Stroke   & 24 / 50 & 10--20 & 30 & 500 Hz & 40 & 1000 \\
2019-Stroke & 15 / 15 & 10--10 & 63 & 512 Hz & 80 & 1708 \\
\bottomrule
\end{tabular}
\end{table}

\subsection{Implementation Details and Hyperparameter Settings}
The proposed framework was developed using PyTorch in a Python 3.12 environment. Optimization was performed with Adam, where both the initial learning rate and weight decay were set to 0.001. Each training run was allowed to proceed for a maximum of 200 epochs. All experiments were executed on a workstation equipped with an AMD EPYC processor and an NVIDIA RTX 3090 GPU. The architectural hyperparameters and dataset-specific adaptation settings selected after parameter testing are jointly summarized in Table~\ref{tab:model-hparams}.

\begin{table}[!t]
\caption{Selected CFSPMNet hyperparameters.}
\label{tab:model-hparams}
\centering
\small
\begin{tabular}{l l l}
\toprule
\textbf{Parameter} & \textbf{XW-Stroke} & \textbf{2019-Stroke} \\
\midrule
Embedding size $d_{\mathrm{emo}}$ & 30 & 38 \\
Encoder depth $L$ & 2 & 2 \\
Temporal filters $F_t$ & 8 & 8 \\
Source loss weight $\alpha$ & 0.98 & 0.95 \\
Pseudo-label threshold $\tau_p$ & 0.60 & 0.60 \\
Fourier partition ratio $r_{\mathrm{spec}}$ & 0.45 & 0.50 \\
Fourier sparsity threshold $\tau_f$ & 0.01 & 0.01 \\
Stage-I epochs $E_{\mathrm{I}}$ & 25 & 10 \\
\bottomrule
\end{tabular}
\end{table}

\subsection{Evaluation Protocol}
Model evaluation followed a leave-one-subject-out (LOSO) protocol. For each LOSO split, one stroke patient was held out as the target subject, while the remaining patients were used as the source domain. The model was trained using source-domain samples and then tested on the unseen target subject. This process was repeated sequentially until each subject had served once as the target domain. To prevent data leakage, the subject-wise partition was completed before model training, ensuring that labeled data from the target subject were never used during source-domain training and that no label information was transferred between the source and target domains.

The reported results were computed by averaging the evaluation metrics over all LOSO rounds. Cross-subject generalization performance was assessed using classification accuracy, Cohen’s kappa coefficient, recall, precision, and F1-score. Statistical comparisons between competing methods were further conducted using the Wilcoxon signed-rank test to obtain the corresponding p-values.

\section{Results and Discussion}
\label{sec:results}

\subsection{Experimental Results}

To evaluate the cross-patient decoding performance of CFSPMNet in post-stroke MI-EEG, we conducted comprehensive comparisons against various state-of-the-art baselines using the LOSO protocol. The selected baselines encompass a wide range of deep learning architectures and adaptation strategies: (1) Convolutional Neural Network (CNN)-based models, specifically EEGNet\cite{Lawhern2018}, IFNet\cite{Wang2023IFNet}, and ShallowConvNet\cite{Schirrmeister2017}; (2) Transformer-based models, including DBConformer\cite{Wang2025DBConformer}, EEGConformer\cite{Song2023EEGConformer}, and MSCFormer\cite{Zhao2025MSCFormer}; (3) a Mamba-based model, SlimSeiz\cite{Lu2024SlimSeiz}; and (4) transfer learning or domain adaptation methods, such as SSTDA\cite{Chen2025SSTDA}, UA-DANN\cite{Shen2025UADAAN}, and SSAS\cite{Liu2026SSAS}.

As shown in Tables~\ref{tab:overall_xw} and~\ref{tab:overall_2019}, CFSPMNet achieved the best average performance on both stroke datasets. On XW-Stroke, CFSPMNet demonstrated an accuracy of 68.23\%, marking a substantial improvement of 5.63 percentage points over the strongest competitor, SSTDA. On 2019-Stroke, CFSPMNet reached 73.33\% accuracy, exceeding the strongest competing method, SSAS, by 8.25 percentage points. CFSPMNet also achieved the highest F1-score and kappa on both datasets, which robustly confirms its effectiveness in achieving balanced class discrimination and agreement beyond chance.

This observed performance pattern provides critical insights into stroke-patient MI-EEG decoding. While CNN and Transformer baselines can extract local or global EEG dependencies, they often fall short in post-stroke MI-EEG requiring trial-level reorganization of disrupted latent neural-state arrangements. The advantage over SlimSeiz suggests that state-space propagation benefits from explicit Fourier-derived trial context. The improvement over adaptation baselines further indicates that target-domain pseudo-labeling becomes more reliable when semantic confidence is constrained by shared-private physiological consistency. On 2019-Stroke, the standard deviation remains large, likely reflecting pronounced inter-patient variability, yet CFSPMNet still maintains the highest mean values across all reported metrics. On XW-Stroke, all baseline comparisons reached $p<0.01$, supporting a stable advantage under the selected stroke-patient cohort.

\begin{table}[!t]
\caption{MI classification performance on XW-Stroke.}
\label{tab:overall_xw}
\centering
\scriptsize
\renewcommand{\arraystretch}{1.1}
\begin{tabular*}{\textwidth}{@{\extracolsep{\fill}}lccccc@{}}
\toprule
\textbf{Method} & \textbf{Accuracy (\%)} & \textbf{Recall (\%)} & \textbf{Precision (\%)} & \textbf{F1-Score (\%)} & \textbf{Kappa} \\
\midrule
EEGNet\cite{Lawhern2018}$^{**}$          & 61.56 ± 06.53 & 61.87 ± 13.83 & 62.00 ± 07.90 & 61.11 ± 08.37 & 0.231 ± 0.131 \\
IFNet\cite{Wang2023IFNet}$^{**}$                    & 61.88 ± 07.15 & 54.38 ± 17.22 & 67.20 ± 11.94 & 57.23 ± 13.25 & 0.237 ± 0.143 \\
ShallowConvNet\cite{Schirrmeister2017}$^{**}$     & 62.29 ± 05.44 & 65.42 ± 14.14 & 62.32 ± 05.92 & 62.68 ± 08.57 & 0.246 ± 0.109 \\
DBConformer\cite{Wang2025DBConformer}$^{**}$      & 61.15 ± 04.89 & 62.29 ± 26.34 & 67.49 ± 14.54 & 58.45 ± 14.19 & 0.223 ± 0.098 \\
EEGConformer\cite{Song2023EEGConformer}$^{**}$    & 61.67 ± 04.66 & 65.00 ± 17.38 & 62.83 ± 06.48 & 61.68 ± 09.43 & 0.233 ± 0.093 \\
MSCFormer\cite{Zhao2025MSCFormer}$^{**}$          & 58.96 ± 05.30 & 61.88 ± 26.88 & 61.87 ± 10.33 & 56.89 ± 14.89 & 0.179 ± 0.106 \\
SlimSeiz\cite{Lu2024SlimSeiz}$^{**}$              & 58.75 ± 05.64 & 53.12 ± 22.68 & 63.41 ± 11.21 & 53.70 ± 13.62 & 0.175 ± 0.113 \\
SSTDA\cite{Chen2025SSTDA}$^{**}$                  & 62.60 ± 05.57 & 54.79 ± 18.00 & 67.45 ± 11.04 & 57.98 ± 10.33 & 0.252 ± 0.111 \\
UA-DANN\cite{Shen2025UADAAN}$^{**}$               & 58.33 ± 03.86 & 49.17 ± 15.72 & 60.68 ± 04.52 & 52.81 ± 09.59 & 0.167 ± 0.077 \\
SSAS\cite{Liu2026SSAS}$^{**}$                     & 60.42 ± 06.15 & 57.08 ± 18.59 & 59.59 ± 14.19 & 56.98 ± 15.06 & 0.208 ± 0.123 \\
\textbf{CFSPMNet} & \textbf{68.23 ± 05.13} & \textbf{67.08 ± 16.83} & \textbf{69.91 ± 05.93} & \textbf{66.78 ± 09.29} & \textbf{0.365 ± 0.103} \\
\bottomrule
\end{tabular*}
\vspace{2pt}
\raggedright
\footnotesize{$^{*}$ and $^{**}$ indicate statistically significant differences in Accuracy compared with CFSPMNet at $p<0.05$ and $p<0.01$, respectively.}
\end{table}

\begin{table}[!t]
\caption{MI classification performance on 2019-Stroke.}
\label{tab:overall_2019}
\centering
\scriptsize
\renewcommand{\arraystretch}{1.1}
\begin{tabular*}{\textwidth}{@{\extracolsep{\fill}}lccccc@{}}
\toprule
\textbf{Method} & \textbf{Accuracy (\%)} & \textbf{Recall (\%)} & \textbf{Precision (\%)} & \textbf{F1-Score (\%)} & \textbf{Kappa} \\
\midrule
EEGNet\cite{Lawhern2018}$^{*}$                          & 55.83 ± 06.61 & 50.67 ± 31.94 & 61.09 ± 23.42 & 46.35 ± 24.93 & 0.117 ± 0.132 \\
IFNet\cite{Wang2023IFNet}$^{**}$                         & 55.42 ± 04.44 & 52.00 ± 22.01 & 60.62 ± 12.74 & 51.21 ± 13.35 & 0.108 ± 0.089 \\
ShallowConvNet\cite{Schirrmeister2017}$^{**}$            & 57.25 ± 07.43 & 54.67 ± 37.06 & 58.06 ± 27.49 & 47.84 ± 26.24 & 0.145 ± 0.149 \\
DBConformer\cite{Wang2025DBConformer}$^{**}$             & 56.17 ± 07.90 & 43.00 ± 39.45 & 63.68 ± 30.54 & 38.63 ± 28.89 & 0.123 ± 0.158 \\
EEGConformer\cite{Song2023EEGConformer}$^{*}$          & 58.17 ± 04.81 & 58.17 ± 26.39 & 63.73 ± 12.67 & 54.67 ± 16.02 & 0.163 ± 0.096 \\
MSCFormer\cite{Zhao2025MSCFormer}$^{*}$                & 60.08 ± 08.07 & 57.00 ± 35.45 & 66.58 ± 24.90 & 52.05 ± 23.17 & 0.202 ± 0.161 \\
SlimSeiz\cite{Lu2024SlimSeiz}$^{*}$                    & 57.33 ± 07.50 & 47.50 ± 32.44 & 66.70 ± 15.35 & 46.09 ± 23.24 & 0.147 ± 0.150 \\
SSTDA\cite{Chen2025SSTDA}                       & 63.50 ± 15.99 & 53.83 ± 35.61 & 58.78 ± 26.91 & 52.20 ± 31.10 & 0.270 ± 0.320 \\
UA-DANN\cite{Shen2025UADAAN}                    & 61.75 ± 05.45 & 68.33 ± 18.23 & 62.11 ± 06.61 & 62.86 ± 09.94 & 0.235 ± 0.109 \\
SSAS\cite{Liu2026SSAS}                          & 65.08 ± 20.12 & 68.33 ± 26.72 & 64.27 ± 20.91 & 64.59 ± 23.55 & 0.302 ± 0.402 \\
\textbf{CFSPMNet} & \textbf{73.33 ± 18.71} & \textbf{81.50 ± 21.21} & \textbf{73.28 ± 19.59} & \textbf{75.11 ± 19.38} & \textbf{0.467 ± 0.374} \\
\bottomrule
\end{tabular*}
\vspace{2pt}
\raggedright
\footnotesize{$^{*}$ and $^{**}$ indicate statistically significant differences in Accuracy compared with CFSPMNet at $p<0.05$ and $p<0.01$, respectively.}
\end{table}

\subsection{Ablation Study}

To analyze the contribution of each component, we performed module-level ablation studies on both datasets. As summarized in Table~\ref{tab:ablation}, every ablated variant reduced accuracy compared with the full CFSPMNet, indicating that the two proposed modules provide complementary benefits for cross-patient post-stroke MI-EEG decoding.

Removing the SPPM strategy resulted in a decrease in accuracy of 2.33 percentage points on 2019-Stroke and 4.27 percentage points on XW-Stroke. This result indicates that target-sample selection requires two simultaneous criteria: semantic confidence and compatibility between private physiological signatures and shared class prototypes learned from source patients. In post-stroke MI-EEG, altered cortical excitability and network coupling can produce confident predictions whose latent state organization deviates from the shared class structure. SPPM therefore acts as a physiological acceptance boundary: shared prototypes retain cross-patient class anchors, while private signatures preserve target-patient state characteristics. Furthermore, the variant without dynamic pseudo-label updating incurred even larger performance drops, particularly a substantial 8.58 percentage points on 2019-Stroke. This emphatically demonstrates that target supervision must dynamically adapt to the evolving target representation, preventing the model from being locked into patient-specific ambiguities early in the adaptation process. Dynamic updating allows pseudo-supervision to be continuously refreshed, aligning with the progressive refinement of the target-patient's neural-state organization.

The FRSM-related ablations further demonstrate the physiological contribution of Fourier-domain token-state reorganization. Removing Fourier context guidance reduced accuracy by 4.08 percentage points on 2019-Stroke and 4.48 percentage points on XW-Stroke, showing that Mamba state propagation benefits from trial-specific Fourier-derived context. From a neural-state perspective, the Fourier-derived context provides a compact description of how latent physiological tokens are coordinated within the current trial. This is important for stroke patients because changes in aperiodic operating points, cortical excitability, and inter-areal coordination can alter the ordering, coupling, and amplitude-phase coordination of class-related states even for the same MI class. Removing the High-Frequency Branch or Low-Frequency Branch also degraded performance, with drops of 4.33 and 6.08 percentage points on 2019-Stroke and 4.69 and 5.31 percentage points on XW-Stroke. These results indicate that the two Fourier branches provide non-redundant views of the trial-level physiological organization and jointly support context-conditioned state-space propagation.

Taken together, the ablation results support the synergistic design of CFSPMNet. FRSM improves the organization of within-trial latent neural states before state propagation, while SPPM constrains target-domain supervision using shared-private physiological consistency. The full model therefore links physiological token-state reorganization with calibrated pseudo-label updating, which is useful when the decoding substrate is reshaped at local, cross-regional, and global state levels.

\begin{table}[!t]
\caption{Ablation results across datasets.}
\label{tab:ablation}
\centering
\scriptsize
\resizebox{0.65\columnwidth}{!}{%
\begin{tabular}{llll}
\toprule
\textbf{Dataset} & \textbf{Variant} & \textbf{Accuracy (\%)} & \textbf{$\Delta$Acc. (\%)} \\
\midrule
\multirow{6}{*}{2019-Stroke}
& \textbf{CFSPMNet} &     \textbf{73.33 ± 18.71} & -- \\
& w/o SPPM Strategy             & 71.00 ± 17.54 & -2.33 \\
& w/o Dynamic Pseudo Update     & 64.75 ± 13.59 & -8.58 \\
& w/o Fourier Context Guidance  & 69.25 ± 13.10 & -4.08 \\
& w/o High-Frequency Branch & 69.00 ± 17.52 & -4.33 \\
& w/o Low-Frequency Branch & 67.25 ± 17.41 & -6.08 \\
\midrule
\multirow{6}{*}{XW-Stroke}
& \textbf{CFSPMNet} &     \textbf{68.23 ± 5.13} & -- \\
& w/o SPPM Strategy             & 63.96 ± 6.96  & -4.27 \\
& w/o Dynamic Pseudo Update     & 63.65 ± 6.57 & -4.58 \\
& w/o Fourier Context Guidance  & 63.75 ± 5.73 & -4.48 \\
& w/o High-Frequency Branch & 63.54 ± 7.74 & -4.69 \\
& w/o Low-Frequency Branch & 62.92 ± 5.09 & -5.31 \\
\midrule
\end{tabular}
}
\end{table}

\subsection{Hyperparameter Sensitivity Analysis}

\begin{figure}[!t]
\centering
\includegraphics[width=\columnwidth]{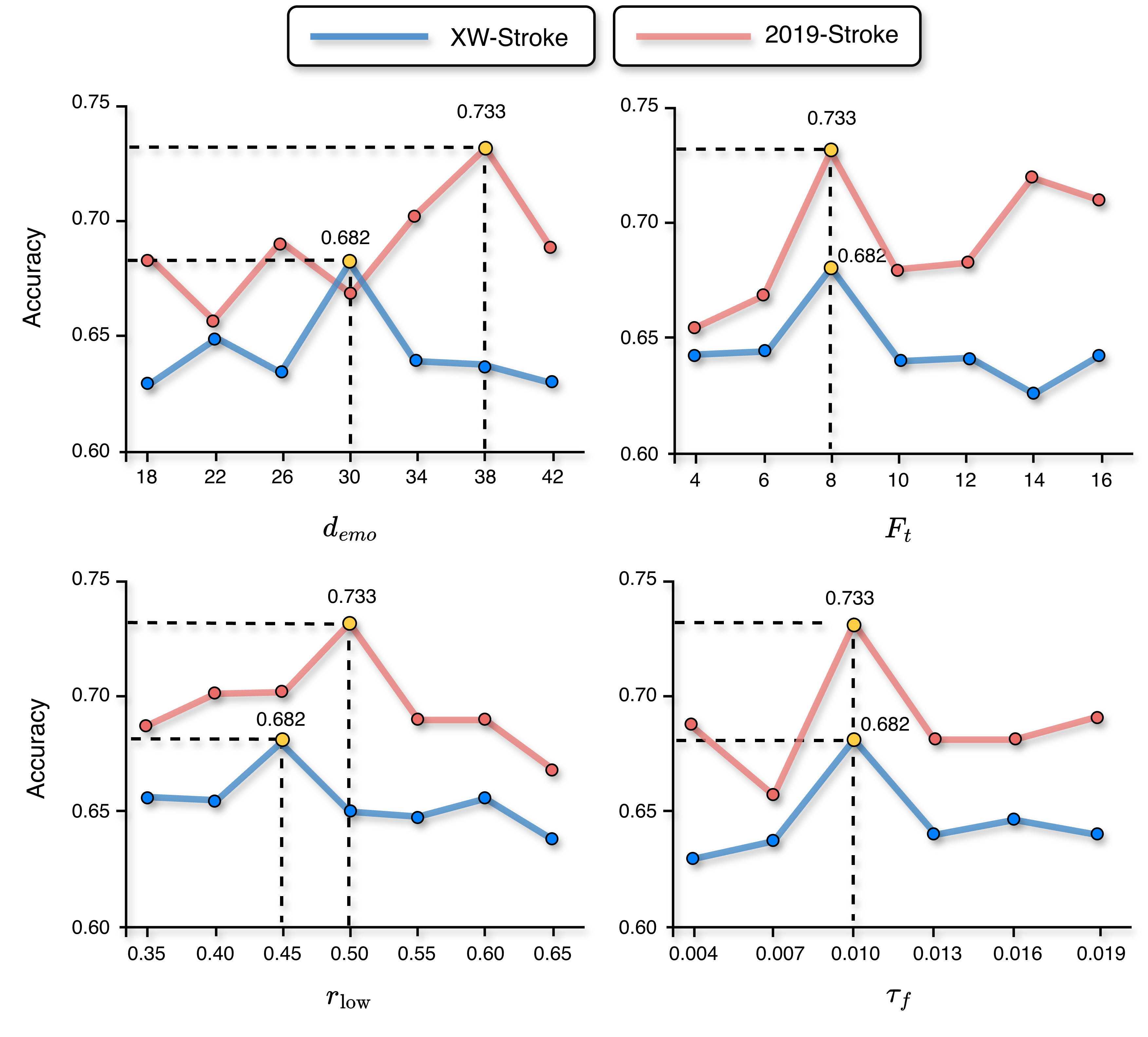}
\caption{Hyperparameter sensitivity on XW-Stroke and 2019-Stroke.}
\label{fig:hyper}
\end{figure}

To evaluate sensitivity to key architectural and adaptation parameters, we tested the embedding size $d_{\mathrm{emo}}$, temporal filters $F_t$, Fourier partition ratio $r_{\mathrm{spec}}$, and Fourier sparsity threshold $\tau_f$. As shown in Fig.~\ref{fig:hyper}, the best embedding sizes were $30$ for XW-Stroke and $38$ for 2019-Stroke, while the best temporal filter number was $8$ on both datasets. This pattern indicates that the latent token capacity should match the inherent complexity of each dataset (e.g., varying neurological heterogeneity or signal characteristics), whereas a moderate temporal front-end is sufficient before Fourier-domain reorganization.

The Fourier parameters also showed clear optima: $r_{\mathrm{spec}}$ was $0.45$ on XW-Stroke and $0.50$ on 2019-Stroke, while $\tau_f$ was $0.01$ on both datasets. Performance decreased when the partition ratio or sparsity threshold moved away from these values, suggesting that complementary Fourier states must preserve stable trial-level organization without retaining excessive perturbations. Parameter testing further showed that the remaining settings in Table~\ref{tab:model-hparams} were optimal under the evaluated grid and were fixed when plotting Fig.~\ref{fig:hyper}: $L$ was $2$, $\alpha$ was $0.98/0.95$, $\tau_p$ was $0.60$, and $E_{\mathrm{I}}$ was $25/10$ for XW-Stroke/2019-Stroke. Overall, CFSPMNet performed best under a balanced configuration of representation capacity, Fourier-state reorganization, source anchoring, and calibrated target adaptation.

\subsection{Visualization of Feature Alignment}

\begin{figure}[!t]
\centering
\includegraphics[width=\columnwidth]{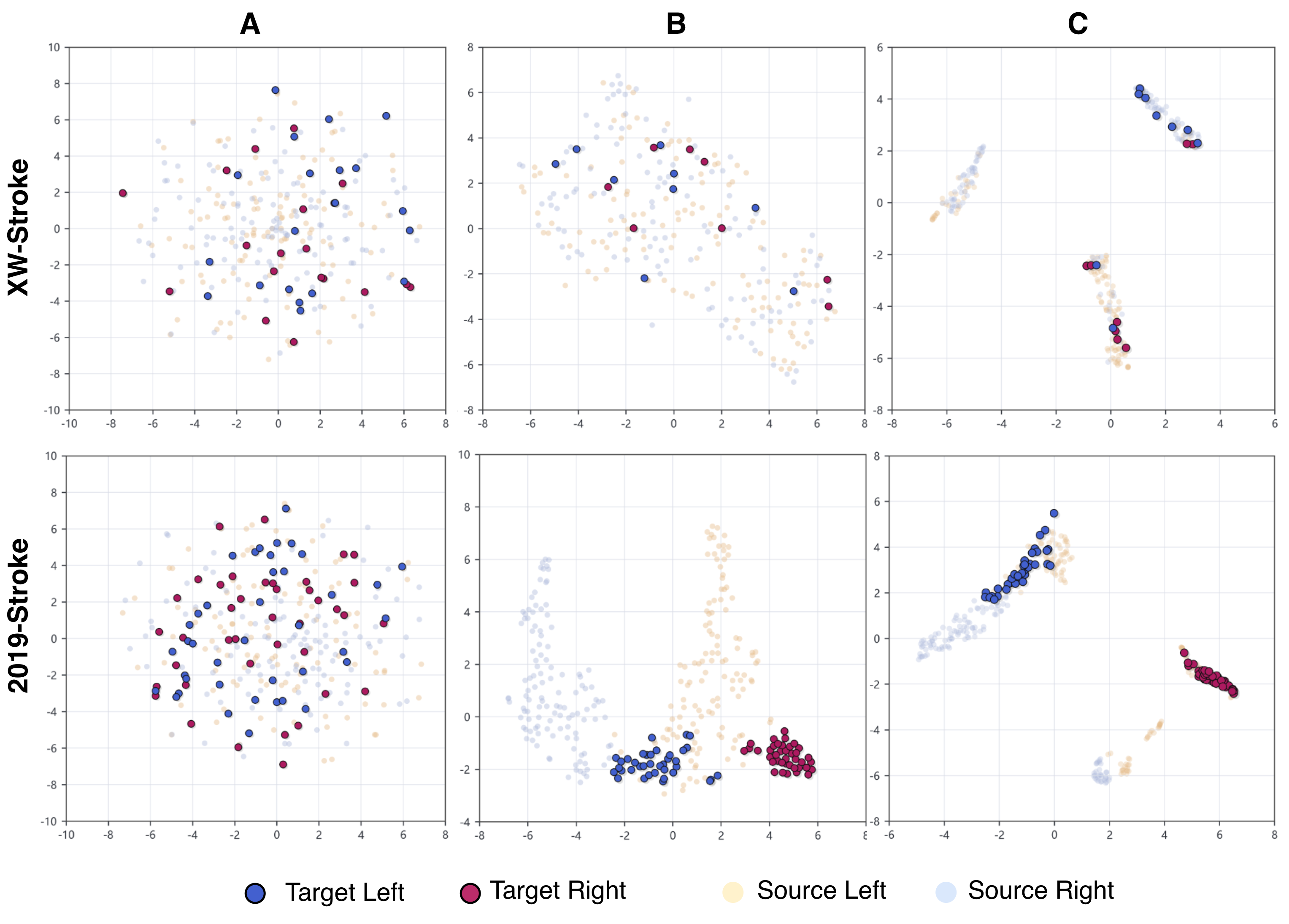}
\caption{t-SNE feature distributions on XW-Stroke and 2019-Stroke.}
\label{fig:tsne}
\end{figure}

To qualitatively assess the effectiveness of CFSPMNet in feature space transformation, we visualize the feature distributions using t-SNE\cite{Maaten2008} in Fig.~\ref{fig:tsne}. Fig.~\ref{fig:tsne}(A) illustrates the initial representation space, where both source and target samples are broadly scattered, and the two MI classes exhibit weak organization. This indicates that direct cross-patient transfer is affected by patient-specific latent state mismatches. Fig.~\ref{fig:tsne}(B) shows the distributions after source-supervised representation learning and intermediate adaptation, where class-related structures become more discernible, yet a notable portion of target samples still cluster ambiguously with source regions. This observation aligns perfectly with the inherent difficulty of post-stroke MI-EEG decoding, where identical MI labels can correspond to various trial-level state trajectories across patients.

Fig.~\ref{fig:tsne}(C) reveals the final CFSPMNet representation space, where target samples form significantly more compact and well-separated class-specific groups, achieving superior alignment with the corresponding source-domain structures on both datasets. The left-hand and right-hand MI samples are also more clearly separated, especially after the shared-private pseudo-label calibration stage. This progression suggests that Fourier-domain token-state reorganization improves the ordering and coordination of trial-level latent physiological tokens, while SPPM further suppresses target samples whose semantic prediction is inconsistent with the shared physiological class structure. The visualization therefore supports the proposed design: CFSPMNet reorganizes cross-patient neural-state representations into a more discriminative and physiologically consistent feature space, beyond simple statistical discrepancy reduction to address underlying neurophysiological variations.

\subsection{Visualization of Fourier-Domain Token-State Reorganization}

\begin{figure}[!t]
\centering
\includegraphics[width=\columnwidth]{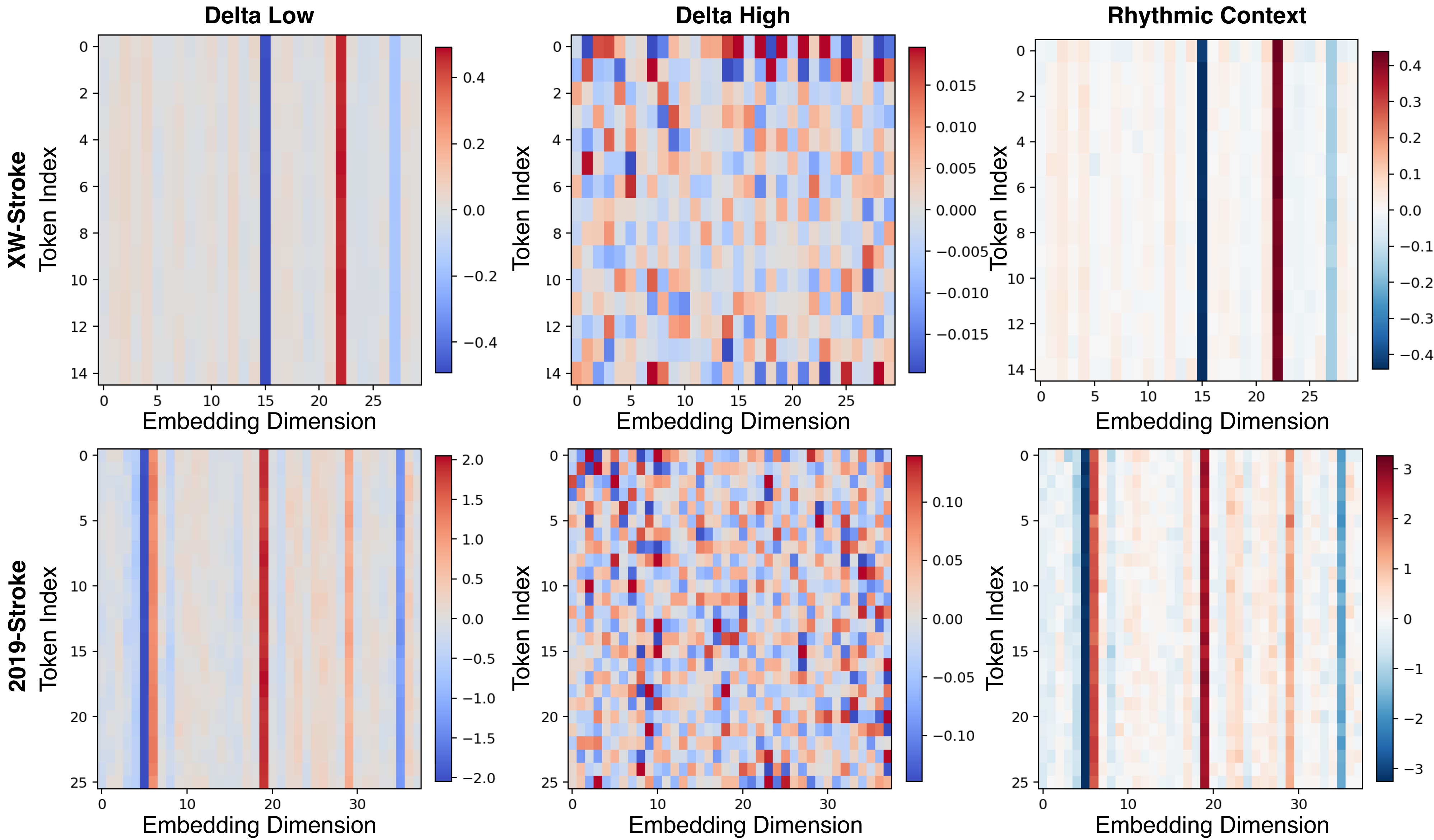}
\caption{Fourier-domain token-state reorganization on XW-Stroke and 2019-Stroke.}
\label{fig:block}
\end{figure}

To further examine the physiological role of FRSM, Fig.~\ref{fig:block} visualizes the block-level token modulation and the resulting Fourier-derived context on the two stroke datasets. In post-stroke MI-EEG, trial-to-trial states are shaped by background cortical excitability, aperiodic electrophysiological activity, and network-level coordination after injury \cite{https://doi.org/10.1113/EP093171, 10.1523/JNEUROSCI.1041-25.2025}. Therefore, an effective rehabilitation-oriented decoder should preserve trial-specific physiological organization before sequential state propagation.

As shown in Fig.~\ref{fig:block}, the branch-specific modulation maps contain structured variations across token index and embedding dimension, while the Fourier-derived context exhibits more coherent and concentrated organization. This significant difference suggests that FRSM does more than amplify isolated local fluctuations. It reorganizes the latent physiological tokens into a trial-level state arrangement that can condition the subsequent Mamba propagation process. Such context-conditioned propagation is relevant for stroke patients because their unique lesion conditions and compensatory recruitment mechanisms fundamentally alter how motor-imagery-related states are coordinated within each individual trial \cite{10.1038/s41540-025-00626-7}. The visualization therefore supports the interpretation that the Fourier-domain operation provides a physiologically informed organizational basis for state propagation, helping the model retain motor-intention cues that are useful for cross-patient rehabilitation decoding.

\subsection{Analysis of Calibrated Target Pseudo-Label Selection}

\begin{figure}[!t]
\centering
\includegraphics[width=\columnwidth]{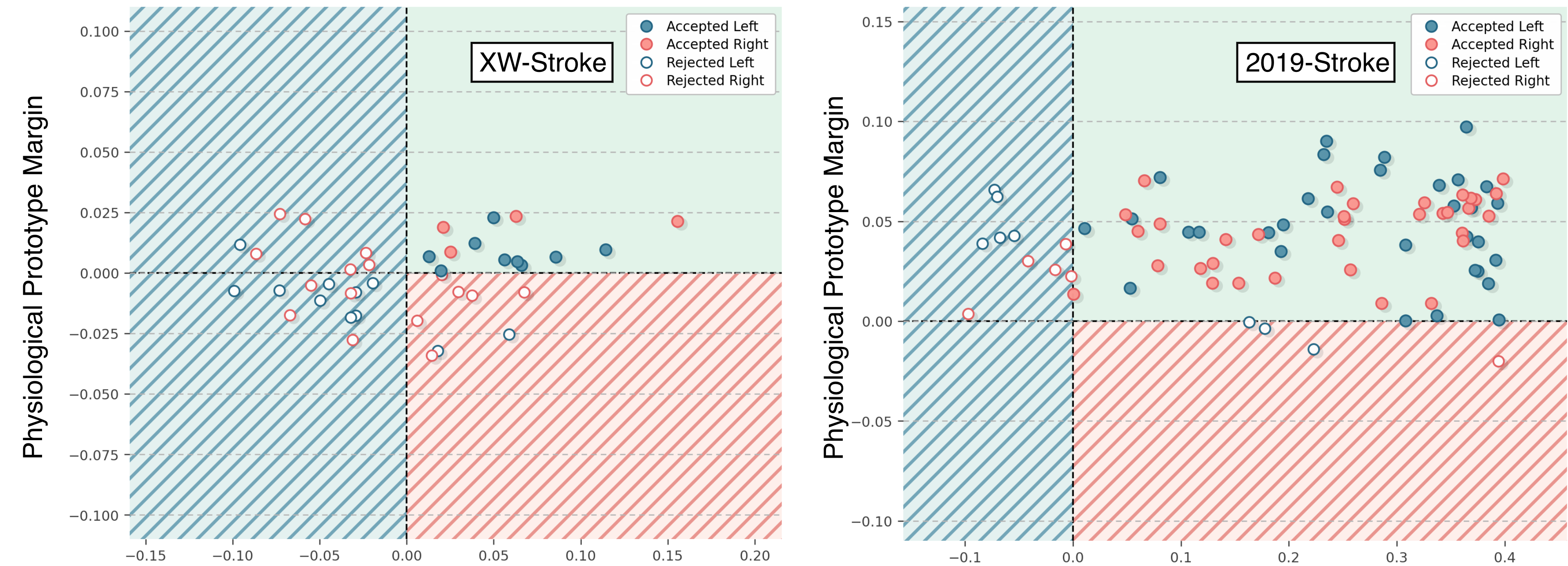}
\caption{Calibrated target pseudo-label selection on XW-Stroke and 2019-Stroke.}
\label{fig:guide_acc}
\end{figure}

Fig.~\ref{fig:guide_acc} illustrates how target-domain samples are accepted or rejected during calibrated pseudo-label updating. For rehabilitation-oriented BCI systems, unreliable pseudo-labels can reinforce incorrect motor-intention feedback for the target patient \cite{https://doi.org/10.1016/j.eswa.2025.127312}. This risk is more pronounced in post-stroke MI-EEG, where a confident prediction may still be pathologically inconsistent with the shared physiological class organization because the patient's latent neural-state trajectory can be reshaped by altered cortical excitability and network coordination.

In both datasets, accepted target samples are mainly distributed in the region where the semantic confidence margin and the physiological-consistency margin are simultaneously positive. Conversely, rejected samples appear in regions where at least one margin is insufficient, including candidates whose predicted class may be confident but whose physiological consistency fails to satisfy the shared-private acceptance criterion. This behavior indicates that SPPM does not rely on prediction confidence alone \cite{10.1109/JBHI.2025.3595826}. Instead, it constrains target-domain supervision by requiring the predicted motor-imagery label to be multi-dimensionally compatible with both global shared class anchors and the patient's individual private physiological signature. From a rehabilitation perspective, this calibrated updating mechanism is important because adaptive decoding should strengthen target-patient supervision only when the candidate label is consistent with a plausible neural-state organization \cite{LIN2026115150}.

\subsection{Spatial Neurophysiological Interpretability Analysis}

\begin{figure}[!t]
\centering
\includegraphics[width=\columnwidth]{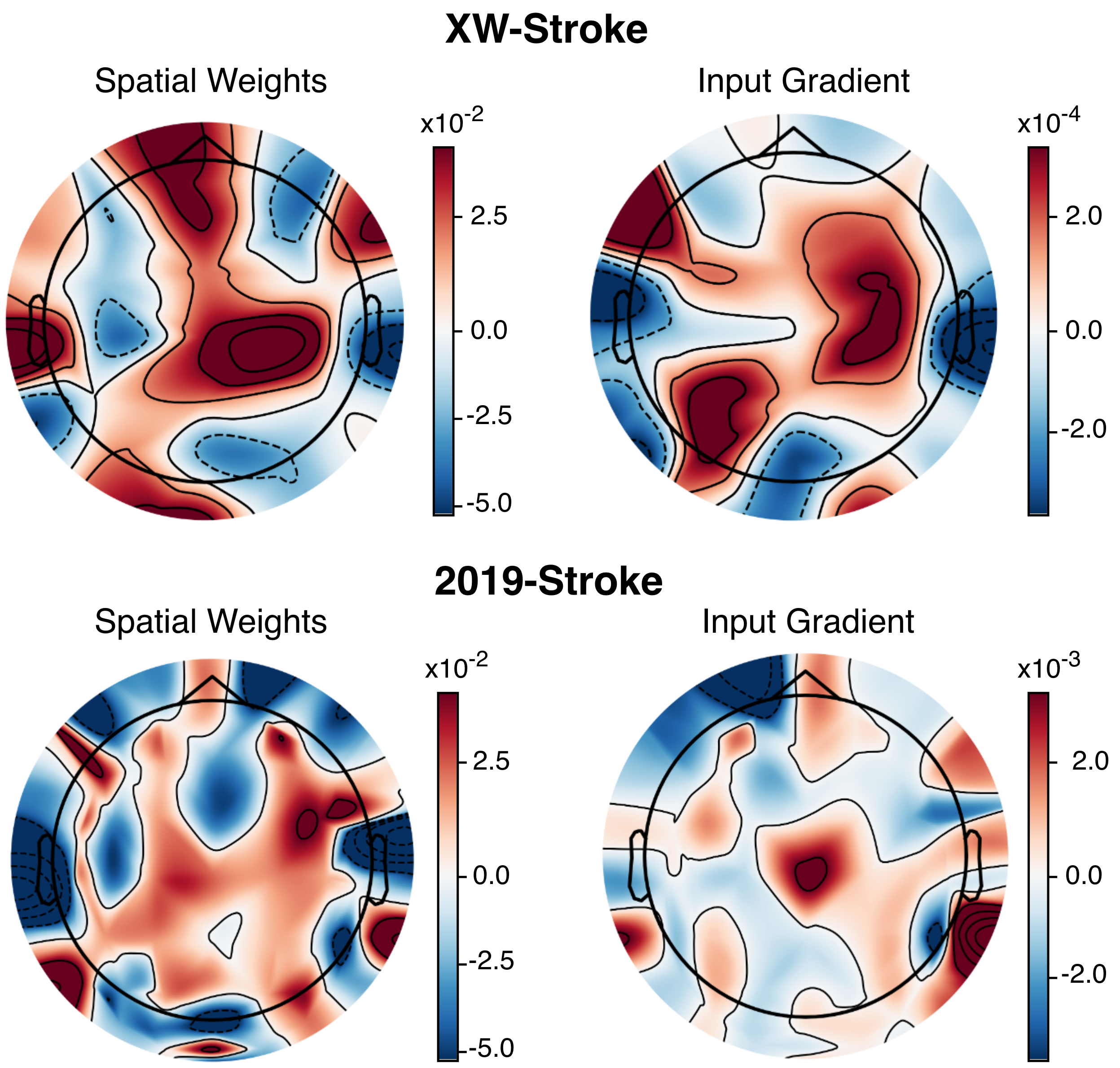}
\caption{Spatial neurophysiological interpretability on XW-Stroke and 2019-Stroke.}
\label{fig:topomap}
\end{figure}

To assess whether the learned representation exhibits physiologically meaningful spatial organization, Fig.~\ref{fig:topomap} presents scalp-level spatial weights and input-gradient responses on both stroke datasets \cite{10.1186/s12984-025-01736-3}. The spatial weights summarize the channel organization learned by the spatial filtering front-end, whereas the input gradients estimate how sensitive the decoder output is to perturbations in the input EEG. These maps are interpreted as scalp-level model-sensitivity summaries without implying precise anatomical localization.

Across the two datasets, we observe distinct non-uniform spatial distributions in both the spatial weights and input-gradient maps. This strongly suggests that CFSPMNet leverages organized spatial patterns with channel-dependent sensitivity rather than treating all channels equally. The spatial patterns also differ between XW-Stroke and 2019-Stroke, which is consistent with evidence that post-stroke motor-state information can involve distributed sensorimotor circuits and recovery-related network interactions \cite{10.1177/10538135261441843}. The presence of distributed and dataset-dependent sensitivity reinforces the central motivation of this study that effective cross-patient stroke MI-EEG decoding fundamentally requires accounting for disrupted latent neural-state organization and patient-specific physiological signatures. For rehabilitation applications, such spatially organized sensitivity provides additional evidence that the model's decisions are linked to neural-system structure relevant to motor-intention decoding.

\subsection{Rehabilitation-Oriented Pathophysiological Implications}

The compelling results achieved by CFSPMNet extend far beyond mere classification accuracy, carrying profound implications for clinical post-stroke rehabilitation. For MI-based BCI to be truly effective in rehabilitation, the decoder must reliably link attempted motor intention to a stable and therapeutically actionable neural state that can drive meaningful feedback. FRSM supports this requirement by reorganizing latent physiological tokens before state propagation \cite{Valente2026}, so the sequence model follows the trial-specific state arrangement rather than only local feature responses. SPPM complements this process by accepting target samples only when semantic prediction and shared-private physiological consistency are jointly satisfied, reducing the risk that closed-loop adaptation reinforces confident but misaligned pseudo-supervision. Therefore, CFSPMNet provides a stable cross-patient initialization while retaining target-specific state signatures, supporting a rehabilitation view of post-stroke MI-EEG decoding as neural-state calibration between intention-related brain activity, decoder output, and therapeutic feedback.

\subsection{Limitations and Future Directions}

This study remains an initial exploration of latent-state-organization-based cross-patient decoding. The available public stroke EEG resources are limited, and the XW-Stroke experiments used a clinically screened subcohort; although this improves cohort definition, the complete cohort and broader stroke populations still require evaluation. The relatively large standard deviations also indicate that performance remains sensitive to inter-patient variability, calling for larger and more clinically diverse validation.

The public datasets further lack complete clinical and neuroimaging metadata, which prevents model performance from being stratified according to lesion hemisphere, lesion location, lesion volume, corticospinal tract involvement, disease stage, medication status, rehabilitation history, and quantitative motor-function scores. As a result, the current framework can only adapt from EEG observations, without explicitly modeling how patient-specific pathological factors modulate MI-related rhythms and cross-patient transferability. Future work should therefore move beyond EEG-only adaptation by incorporating structured clinical variables and neuroimaging-derived descriptors into the decoding framework. For example, lesion profiles and motor impairment scores could be used to define clinically meaningful patient subgroups, guide source-patient selection, or regularize pseudo-label updating according to the expected degree of physiological similarity. In addition, prospective multi-center validation should be conducted under harmonized acquisition protocols to examine whether the proposed adaptation mechanism remains stable across hospitals, devices, therapists, and rehabilitation stages. Finally, online pseudo-label updating should be evaluated in repeated rehabilitation sessions, where the key outcome should not be limited to offline decoding accuracy but should also include session-to-session calibration stability, false-feedback reduction, user engagement, and measurable improvement in motor-function recovery. Such analyses would clarify whether improved cross-patient decoding can be translated into more reliable and clinically meaningful therapeutic feedback \cite{he2026,wang2025,11345330}.

\section{Conclusion}
\label{sec:conclusion}

This paper proposes CFSPMNet for cross-patient post-stroke MI-EEG decoding by treating pathological transfer as a latent neural-state organization problem involving task-related dynamics, aperiodic activity, and local-to-global population states. FRSM reorganizes physiological tokens in the Fourier domain and injects trial-specific Fourier-derived context into Mamba state propagation, while SPPM constrains target pseudo-label updating through semantic confidence and shared-private physiological consistency. Under the leave-one-subject-out protocol, CFSPMNet achieved accuracies of 68.23\% on XW-Stroke and 73.33\% on 2019-Stroke, outperforming representative CNN-, Transformer-, Mamba-, and adaptation-based baselines. Ablation, sensitivity, feature-alignment, pseudo-label selection, and spatial interpretability analyses further supported the contributions of Fourier-domain token-state reorganization and calibrated target adaptation. These findings suggest that rehabilitation-oriented post-stroke MI-EEG decoding can benefit from modeling neural-state calibration across patients, and future work should validate this framework in larger multi-center and online rehabilitation settings.

\section*{Reference}
\bibliographystyle{unsrt}
\bibliography{references}

\end{document}